\documentclass[10pt,a4paper]{article}
\usepackage{multicol} %用于实现在同一页中实现不同的分栏
\usepackage{graphicx}
\usepackage{float}
\usepackage{indentfirst} 
\usepackage[utf8]{inputenc} % allow utf-8 input
\usepackage[T1]{fontenc}    % use 8-bit T1 fonts
\usepackage{hyperref}       % hyperlinks
\usepackage{url}            % simple URL typesetting
\usepackage{booktabs}       % professional-quality tables
\usepackage{amsfonts}       % blackboard math symbols
\usepackage{nicefrac}       % compact symbols for 1/2, etc.
\usepackage{microtype}      % microtypography
\usepackage{xcolor}         % colors
\usepackage{times}
\usepackage{epsfig}
\usepackage{graphicx}
\usepackage{amsmath}
\usepackage{multirow}

\usepackage{amssymb}
\usepackage{subcaption}
\usepackage{cleveref}
\usepackage{epstopdf}
\usepackage{array}%需要用到该宏包
\usepackage{footnote}
\makesavenoteenv{tabular}

\usepackage{booktabs}

\usepackage{geometry}
\usepackage{graphicx} % Required for inserting images
\usepackage[runin]{abstract}
\title{Overview of Cross-Component In-loop Filters \\in Video Coding Standards}
\author{Zhaoyu Li, Xuewei Meng, Jiaqi Zhang, Cheng Huang, Chuanmin Jia, Siwei Ma, Yun Jiang}
\date{May 2024}

\begin{document}    

\maketitle

\begin{abstract}
   %{For traditional video coding, this paper summarizes the cross-component filtering tools currently used in video coding. Specifically, it includes the Cross-Component Sample Offset (CCSAO) and Cross-Component Adaptive Loop Filter (CCALF) used in the current mainstream video coding standards. Cross-component technology is an important tool to improve video compression performance by reducing sample distortion through the strong correlation between different components and providing more accurate pixel reconstruction values. Based on this, we report the technical principles and development process of the current mainstream cross-component filtering tools, and give a brief prospect for the development of cross-component filtering.} 
   {
   In-loop filters have been comprehensively explored during the development of video coding standards due to their remarkable noise-reduction capability. In the early stage of video coding, in-loop filters, such as Deblocking Filter, Sample Adaptive Offset, and Adaptive Loop Filter, were performed separately for each component. Recently, cross-component filters were studied to improve the chroma fidelity by exploiting correlations between the luma and chroma channels. This paper summarizes the cross-component filters used in the state-of-the-art video coding standard. Specifically, it includes the Cross-Component Adaptive Loop Filter and Cross-Component Sample Adaptive Offset. Cross-component filters aim to reduce compression artifacts based on the correlation between different components and provide more accurate pixel reconstruction values. In this paper, we introduce the origin, development, and status of cross-component filters in the current video coding standards. Finally, we had some discussions on the further evolutions of cross-component filters.
   }
\end{abstract}

\begin{multicols}{2}       % 分两栏 若花括号中为3则是分三列
{
\section{Introduction}
    {
    
    With the continuous development of video capture, storage, compression, and display technologies, numerous video applications continue to emerge, such as video communications, online conferences, cloud gaming, immersive video experiences, and so on. These advancements bring forth new challenges for video coding technologies. To meet the increasing demand for video compression, various video coding tools and technologies have been proposed, leading to continuous evolution in video coding standards. A significant milestone in this progression was the finalization of the High Efficiency Video Coding (HEVC)~\cite{ref7} standard in 2013, which achieved approximately 50\% bitrate saving compared to its predecessor, the Advanced Video Coding (AVC) standard~\cite{AVC}. The latest video coding standard, Versatile Video Coding (VVC)~\cite{overview}, has further improved upon HEVC by achieving roughly 50\% bitrate reduction. While H.266/VVC demonstrates excellent video compression capabilities, there remains significant potential for enhancing video coding efficiency further.
    In the pursuit of exploring advanced video encoding tools, a software model named the Enhanced Compression Model~(ECM) has been introduced to further explore the potential of video compression~\cite{ECM}. 
    
    %Driven by the adoption of high-definition, ultra-high-definition, and high-dynamic-range videos, there is also a corresponding improvement in video quality. However, 

    As a result of the prevalent utilization of block-based operations and coarse quantization within contemporary video coding standards, artifacts such as blocking and ringing have become inherent in compressed frames, thereby markedly diminishing both objective and subjective quality. To mitigate these compression artifacts, extensive exploration has been conducted on in-loop filter algorithms during the evolution of video coding standards. These filters serve to enhance the quality of reconstructed frames while also furnishing high-fidelity reference frames for subsequent images, thereby facilitating more accurate motion compensation.
    }
    
    \begin{figure*}[htbp]
    \centering
    \includegraphics[width=6.0in]{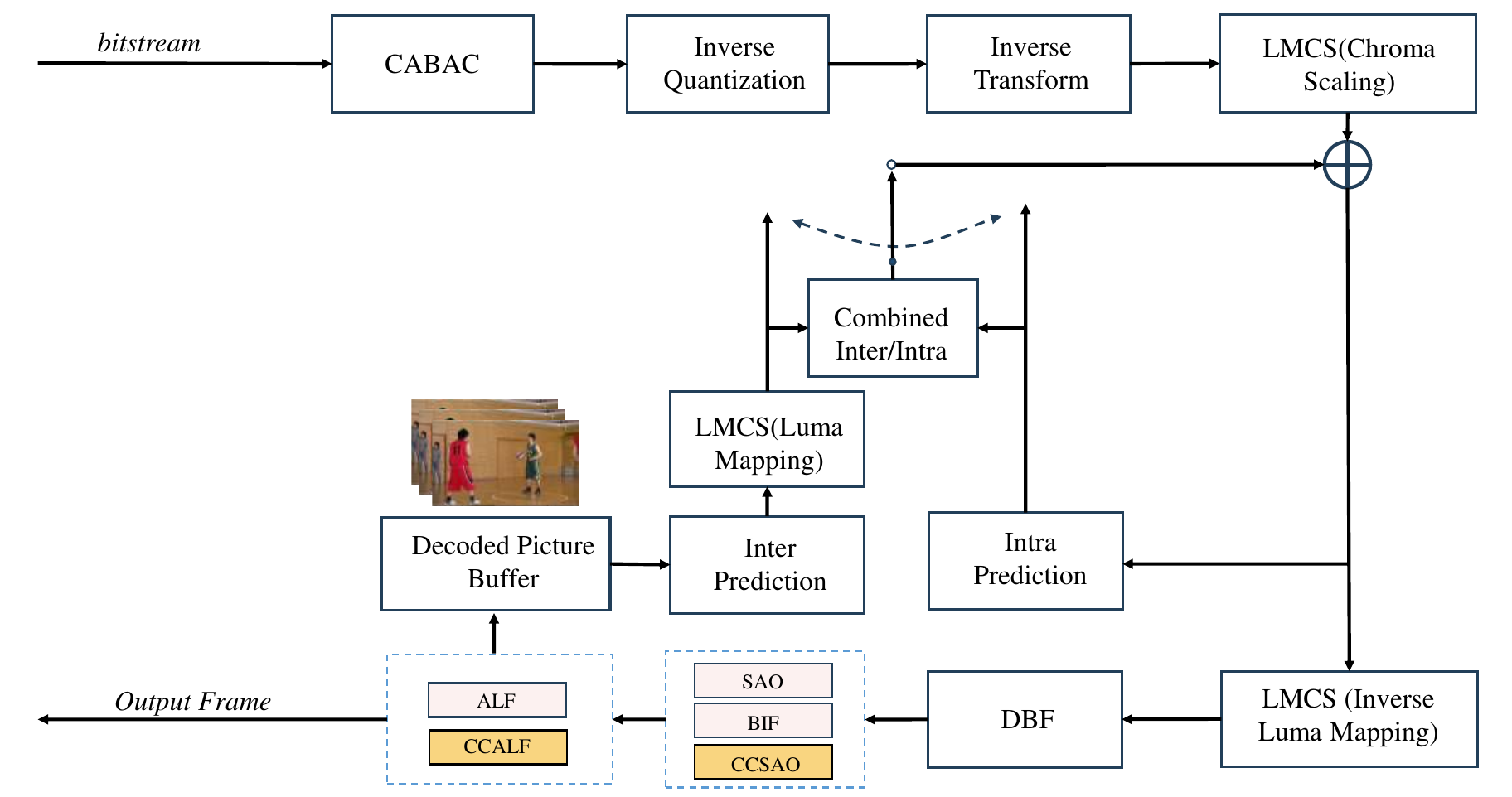} % \textwidth
    \caption{Illustration of the ECM video decoder diagram, with golden boxes corresponding to cross-component filters.}
    \label{fig1}
\end{figure*}
{
    
    There are four kinds of in-loop filters in VVC~\cite{vvcInLoop}, i.e., deblocking filter (DBF)~\cite{ref2}, sample adaptive offset (SAO)~\cite{ref3},  adaptive loop filter (ALF)~\cite{ref1,NonlinearALF-1,NonlinearALF-2,meng2021optimized}, and luma mapping with chroma scaling(LMCS)~\cite{ref4}. The bilateral filter (BIF)\cite{bif}, was newly adopted in ECM. These filters are depicted in Fig.~\ref{fig1}.
    
    %These filters are applied after the picture reconstruction and before the picture in the decoder picture buffer as depicted in Fig.~\ref{fig1}. (TODO: modify this sentence)

    DBF aims to remove the blocking artifact by applying a set of low-pass filters to the boundaries of the coding unit, the prediction unit, and the transform unit. SAO is conducted by conditionally adding an offset to the reconstructed samples after DBF, which shows promising performance in reducing the mean sample distortion and the ringing artifacts. ALF is a Wiener-based spatial filter. It enhances reconstructed video fidelity by taking the weighted average of reference samples as the filtered samples. The weighting coefficients are derived by minimizing the mean square error between the original and decoded samples in the encoder and transmitted to the decoder. LMCS does not particularly focus on artifact reduction but aims to enhance coding efficiency by better utilizing the dynamic range. BIF is a nonlinear, edge-preserving, and noise-reducing filter that is newly introduced in ECM. Similar to the ALF, it also replaces the intensity of each pixel with a weighted average of intensity values from nearby pixels. While, the difference lies in that the weights of BIF depend on the Euclidean distance of pixels and the radiometric differences. This preserves sharp edges. These weights can be calculated both in the encoder and decoder.

}
    {
    %Currently, the primary loop filter tools include non-local filters, neural network-based filters, and loop filtering within traditional coding frameworks as mentioned above., non-local loop filters have garnered attention in the field of video coding and compression
    
    In addition to the above-mentioned local filters adopted in the ECM, some other in-loop filters based on the image non-local similarity have been studied, such as structure-driven adaptive non-local filter(SANF)~\cite{nonlocal3}, non-local structure-based loop filter (NLSF)~\cite{nonlocal2,non-opt,jia2020fast}, novel adaptive loop filter utilizing image non-local prior knowledge~\cite{nonlocal1}, parametric non-local loop filter~(PNLF)~\cite{meng2022parametric}, deformable wiener filter~(DWF)~\cite{dwf}. Some of these methods were also discussed in the JVET meeting~\cite{des,nonProposal1,nonProposal2,nonProposal3}. 
    
    %While these non-local loop filters can yield certain performance gains, their high computational demands and hardware limitations render their application in video coding standards challenging. Therefore, relevant methods to optimize non-local filters have been explored\cite{non-opt,jia2020fast}.

    Though the aforementioned in-loop filters are effective in reducing compression artifacts, these conventional methodologies characterized by hand-craft designs exhibit constraints in addressing more intricate artifacts. In response to this constraint, in-loop filters leveraging convolutional neural networks~(CNNs) have been developed, demonstrating superior performance compared to conventional filtering methods~\cite{nnvc3,nnvc4,nnvc2,nnvc1}. Various neural network-based loop filtering tools have been proposed and adopted by ECM, achieving significant performance improvement~\cite{nnvcproposal1,nnvcproposal2,nnvcproposal3,nnvcproposal4}.

    %At present, the H.266/VVC standard considers elevating video coding compression performance through neural network integration. 
    }
    
    {
    While the coding techniques mentioned above only focus on single-component in-loop filtering, ignoring the correlation between different components. Extensive research has demonstrated that there is a high correlation between luma and chroma components in the YUV format\cite{CC2002,CC2008,CC2009,ccJunruLi}. Based on this correlation, some prediction techniques were proposed, such as Cross Component Prediction (CCP)\cite{ccp} supported in the HEVC Range Extensions and Cross Component Linear Model (CCLM)\cite{cclm}. Besides, cross-component techniques are also applied in end-to-end image compression\cite{ccam}, which effectively improve compression performance. Recently, the correlation between different components was also considered in the in-loop filters.
    }
    
    %{In contemporary conventional compression standards, the selection of the primary image compression format is influenced by various factors such as bit rate, memory utilization, and encoding time. Consequently, YUV420 typically emerges as the preferred choice. In the YUV420 format, there are higher resolution luma components and lower resolution chrome components. Notably, the luma component harbors more intricate data, texture, and structure compared to chroma components. Encoding procedures inherently entail the reduction of redundant data. Thus, leveraging the correlation between luma and chroma components to enhance compression efficacy represents a straightforward and efficient approach. 
    %}

    {
    Several cross-component in-loop techniques were proposed and adopted in H.266/VVC and Audio Video Coding Standard (AVS3), an independently developed Chinese audio-video coding standard. There is also continuous study on these methods during the development of ECM. In the ECM-12.0, there are two cross-component filters, namely Cross Component Adaptive Loop Filter (CCALF) \cite{CC-ALF,CCALF-1,CCALF-2,Parallel,R0225}, and Cross Component Sample Adaptive Offset (CCSAO). CCALF was initially proposed and adopted during the development of H.266/VVC and was optimized and improved in ECM. Similar to ALF, CCALF is also a Wiener Filter. The difference is that it only applies to chroma samples, and it utilizes luma samples as the reference samples and corrects the target chroma pixel by applying a linear filter to these selected luma samples. The filter parameters are trained following the principle of minimizing the mean square error (MSE) in the encoder and transmitted to the decoder. CCSAO was adopted by AVS3 and ECM. Specifically, it uses the correlation between luma and chroma components to classify the reconstructed samples into different categories and assigns each category an offset value for sample adjustment.
    %corrects the reconstructed chroma pixel values by using a linear filter analysis for the luma samples, and the filter parameter value of chroma components is determined through statistical analysis. 
    } 

    {
    %In the traditional coding framework, in-loop filters play an important role due to their remarkable ability in artifact removal. Notably, the cross-component in-loop filtering tools have been adopted as part of the video coding standards, further improving video compression performance. CCALF has been incorporated into H.266/VVC, while CCSAO has been adopted by the third generation of Audio Video coding Standard (AVS3), an independently developed Chinese audio-video coding standard. This article mainly introduces the fundamental principles underlying cross-component in-loop filtering tools within traditional video frameworks. Moreover, this paper primarily introduces the core principles of cross-component in-loop filtering tools within traditional video frameworks and provides a summary and outlook on the development of cross-component filtering tools during the current phase of new-generation video coding standard formulation.
    }
    {
    Compared to ECM-12.0 without CCALF~\cite{ECM}, ECM-12.0 with CCALF achieves 2.49\% and 2.90\% coding gains for the Cb and Cr components under All intra (AI) configuration, and 1.48\% and 2.12\% coding gains for Cb and Cr components under Random Access (RA) configuration. While in the VTM-10.0, CCALF can achieve 13.88\% and 13.73\% coding gains under AI configuration, and 9.69\% and 8.55\% coding gains under RA configuration for Cb and Cr components respectively~\cite{ccalfPerInVVC}. The decrease in the coding gain may caused by the new cross-component techniques introduced in the prediction process of ECM. For CCSAO, 1.28\% and 1.08\% coding gains can be achieved for Cb and Cr components under AI configuration, and 3.02\% and 2.79\% coding gains for Cb and Cr components can be achieved under RA configuration, respectively.
    }
    
    {
    The remainder of this paper is organized as follows, Section 2 introduces the theory of CCALF and summarizes the development of CCALF. Section 3 introduces the fundamental principles of CCSAO and the proposals about CCSAO. Experimental results and discussions are shown in Section 4. Section 5 concludes this paper.
    }
    %Presently, cross-component techniques are extensively integrated into video coding methodologies.
    %and Cross-component adaptive loop filter (CCALF)\cite{ref5}.  Additionally, the ECM has integrated the Cross-Component sample adaptive offset (CCSAO) \cite{ref9}

\section{CCALF}

    CCALF is fundamentally a Wiener Filter \cite{ref5}. Specifically, CCALF derives a correction signal for chroma samples based on the weighted average of luma reference samples. These reference samples are the neighboring samples of the to-be-filtered chroma sample's collaborated luma sample. The collaborated luma sample is derived based on the chroma format of the video. 
    
    %Based on the spatial scale factor between the luma channel and the chroma channel, the luma location is computed, around which the support region is centered. The spatial scaling factor is always determined by the chroma format of a video sequence. 
    
    Both the ALF and the CCALF use the reconstructed sample of SAO as input, while the CCALF only calculates the offsets for chroma components as shown in Fig.~\ref{fig2}. The filtering operation can be represented using the equation below, Let us assume the following for 2-D images.\par
    {\noindent
    1) Sample location $r=(x,y)$ belongs to the to-be-filtered region $R$, $r'=(x',y')$ means the collocated position to the to-be-filtered sample.\\
    2) Original sample: $s[r]$\\
    3) To-be-filtered sample: $I[r]$\\
    4) Collocated luma samples of $I[r]$: $L(r')$\\
    5) $N$-tap filter coefficients: $\mathbf{c} = [c_{0}, c_{1}, c_{2},...,c_{N-1}]$\\
    6) Filter tap position offset: $\left\{\mathbf{d}_{0}, \mathbf{d}_{1}, \mathbf{d}_{2},...,\mathbf{d}_{N-1}\right\}$, where $d_{i}$ denotes the sample location offset to $L(r')$ of the $i$th filter tap.\\
    7) The difference values between neighboring reference luma samples and the to-be-filtered chroma sample: $\mathbf{p} = [p_{0}, p_{1}, p_{2},...,p_{N-1}]$ \\
    8) Filtered chroma sample:$\tilde{I}(r)$
    }

{ \scriptsize
    \begin{equation}    
       % \begin{aligned}
       %    {\tilde{I}(x, y) = I(x, y)+\sum_{i = 0}^{n-1} c_{i} f_{i}+\sum_{i = n}^{n+m-1} c_{i}\left(f_{i, 0}+f_{i, 1}\right)}\\
       %     {+\sum_{i = n+m}^{n+m_{1}-1} c_{i}\left(g_{i, 0}+g_{i, 1}\right)+\sum_{i = n+m_{1}}^{n+m_{j}} c_{i} g_{i},}
       % \end{aligned}
            {\tilde{I}(r) = I(r)+\sum_{i = 0}^{N-1} c_{i} p_{i}},
        \label{eq1}
    \end{equation} 
}

{\noindent
%where $\tilde{I}(x, y)$ is the filtered chroma sample, $(x,y)$ is the coordinate of the to-be-filtered sample, $N$ is the number of coefficients. $f_{i}$ is the difference between reference luma samples and the current chroma sample $I(x, y)$, 
}

{ \scriptsize
    \begin{equation}    
        %\begin{aligned}
           {p_{i}=L(r'+\mathbf{d}_{i})-I(r),}
        %\end{aligned}
        %\label{eq2}
    \end{equation}
}
{\noindent
%where $L(x+x_{i}, y+y_{i})$ represents the reference luma sample
%, and $(x+x_{i}, y+y_{i})$ is the coordinates of the reference luma samples corresponding to coefficient $c_{i}$, 
}
%{\noindent where $f_{i,j}$ and $f_{i}$  is the clipped difference between neighboring samples and current sample $I(x, y)$, $n$ is the number of coefficients in spatial filter that only has one sample corresponding to it, $m$ is the number of coefficients in spatial filter that corresponding to two samples, $g_{i}$  is the value of other input samples of CCALF, $m_{1}-m$ is the number of coefficients in other input filter shapes that corresponding to two sample, at the design of ECM12.0, it is the luma residual samples. $c_{i}$ is the value of corresponding coefficients. $I(x, y)$ is the current sample, and $\tilde{I}(x, y)$ is the value of reconstructed sample value before clipping. }

{
The coefficients of CCALF are derived by minimizing the mean square error between the reconstructed chroma component after SAO and the original chroma sample, similar to the parameter derivation process of Chroma-ALF. Specifically, a correlation matrix is derived, and the coefficients are calculated using the Cholesky decomposition solver to minimize the mean square error. }
{

The coefficient values at different positions are obtained from the bitstream. The filter coefficients are derived by solving the following optimization problem as shown in Eqn.(~\ref{eq3}),
{ \scriptsize
\begin{equation}    
        %\begin{aligned}
           {\min_c \sum_{r\in R}^{} {\left ( \mathbf{c}\odot \mathbf{p} - s\left [ r \right ] \right ) } ^{2} ,}
        %\end{aligned}
        \label{eq3}
    \end{equation}
    \begin{equation}    
        %\begin{aligned}
           {\mathbf{c} = R_{r,r}^{-1} R_{r,s}^{},}
        %\end{aligned}
        \label{eq4}
    \end{equation}
}
{\noindent
where $\odot$ is the inner product. By solving the Wiener-Hopf equations as shown in Eqn.(~\ref{eq4}), the filter coefficients can be calculated. $R_{r,r}^{-1}$ denotes the auto-correlation matrix of the to-be-filtered samples, $R_{r,s}^{}$ is the cross-correlation vector of the to-be-filtered and the original samples.
}
}
\begin{figure}[H]
	\centering
	\includegraphics[width=\columnwidth]{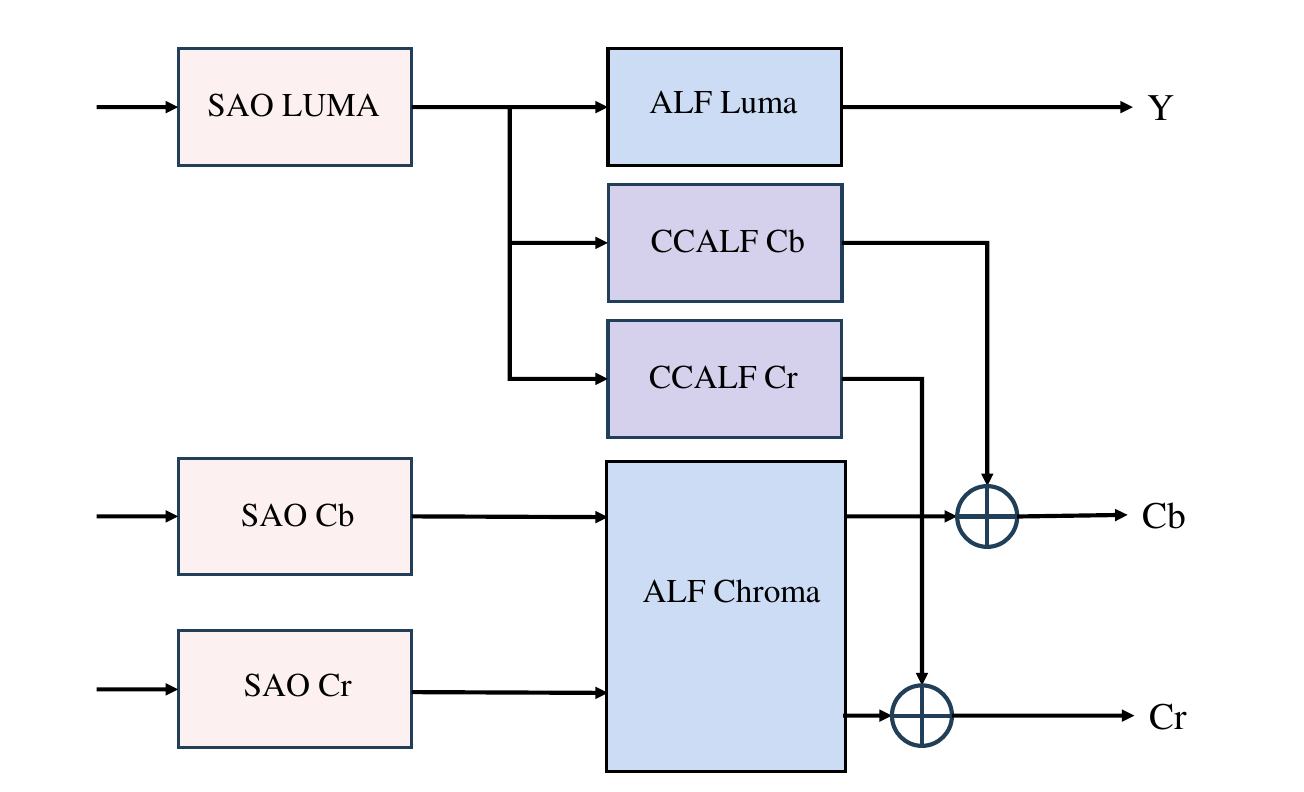} % \textwidth
	\caption{Illustration of CCALF. }
	\label{fig2}
\end{figure}

\if false
\subsection{Evolution of CCALF}
    {   
    CCALF is employed within the in-loop filtering process to enhance chroma fidelity by leveraging the correlation between luma and chroma. In the initial study, ALF was conducted on both luma and chroma components, enabling parallel processing of different sample components \cite{ref1}. However, with the continuous advancement of video coding standards, the robust correlation between the luma and chroma components has been utilized. CCALF was proposed at the 15th JVET meeting in June 2019 and has since been integrated into the H.266/VVC.
    
    CCALF was introduced in H.266/VVC, and the ALF filtering framework containing CCALF is illustrated in Fig.~\ref{fig2}. In H.266/VVC, CCALF utilizes the output luma sample of SAO as input, and its output is employed to refine the chroma reconstructed sample of ALF. Unlike some tools that use luma samples to predict chroma samples, such as the CCLM intra prediction, CCALF doesn't introduce latency as the luma sample used does not require extensive processing before reconstructing the chroma sample. The data flow depicted in Fig.~\ref{fig2} demonstrates that CCALF can be parallelized with ALF.
    }
    {
    In H.266/VVC, the $3\times4$ diamond filter was selected to compute the filter coefficients of CCALF, with the center of the diamond filter aligned with the chroma sample. Fig.~\ref{fig3} illustrates the relative location of the chroma sample being filtered and its support region in the luma sample when CCALF is adopted in H.266/VVC. Consequently, each CCALF filter only needs to store 7 filter coefficients. In H.266/VVC, the CCALF filter coefficients are computed by minimizing the mean square error of each chroma component relative to the original chroma sample. The H.266/VVC algorithm employs a similar parameter derivation process to Chroma-ALF. Specifically, a correlation matrix is derived, and the coefficients are calculated using the Cholesky decomposition solver to minimize the mean square error.
 
    In ALF, the number of coefficients is reduced by utilizing symmetry. However, CCALF does not adopt this method due to its potential impact on compression efficiency. Instead, CCALF employs a more restricted symmetry scheme at the tips of the diamond.
    }

    {
    Due to the abundant texture features of the luma component, CCALF may introduce artifacts with overly abundant chroma texture, thereby reducing the subjective quality of the image, especially at high QP. Therefore, the H.266/VVC reference encoder can achieve subjective tuning by configuring the config file. Specifically, it can attenuate the application of CCALF in high QP encoding and areas with high-frequency luminance. Algorithmically, CCALF is deactivated on CTUs that meet the established conditions.
    }
\begin{figure}[H]
	\centering
	\includegraphics[width=\columnwidth]{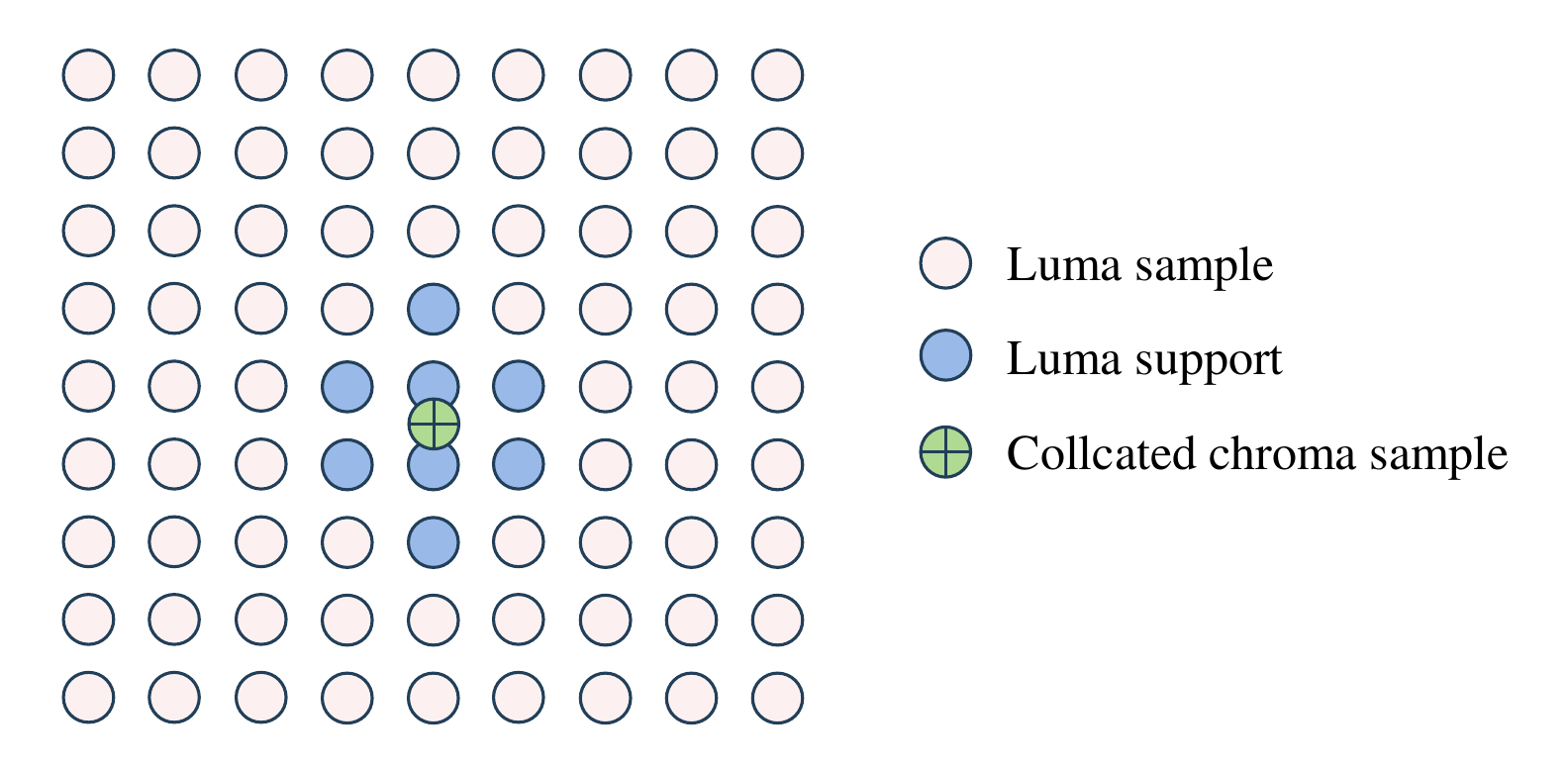} % \textwidth
	\caption{Illustration of the relative location of filtered chroma sample of CCALF and its  support in the luma plane for 4:2:0  chroma format in VVC.}
	\label{fig3}
\end{figure}
%shape of CCALF. The white circles represent the spatial sample, the blue cycles show the luma sample used for chroma center sample, the little green dot represent the collocated chroma sample to be filtered.

    Numerous proposals related to CCALF have been put forward, considering the trade-off between performance and running time, complexity, or other factors. However, some of these proposals were not adopted.

    In \cite{V0080}, an extension to CCALF has been proposed. This contribution suggests extensions to CCALF in both the number and size of filters. While this extension can enhance chroma components, it may lead to some loss in the luma component. Additionally, CCALF could introduce artifacts in chroma components, which is why certain constraints are set in high QP regions. Therefore, proposals regarding CCALF must avoid reintroducing these artifacts.
    
    Considering that the correlation between neighboring pixels may depend on the characteristics of the video content, a single filter shape may not be optimal for all types of video content. To further enhance the performance of CCALF, a Coding Tree Block (CTB) level filter shape selection process has been proposed \cite{ref16}. This contribution introduces two filter shapes, as shown in Fig.~\ref{fig4}. Within each Adaptation Parameter Set (APS), multiple filters and their corresponding shapes with coefficients are signaled. For each CTB, the decoder specifies which filter shape and coefficients are used based on the signaled index. 
    
\begin{figure}[H]
	\centering
	\includegraphics[width=\columnwidth]{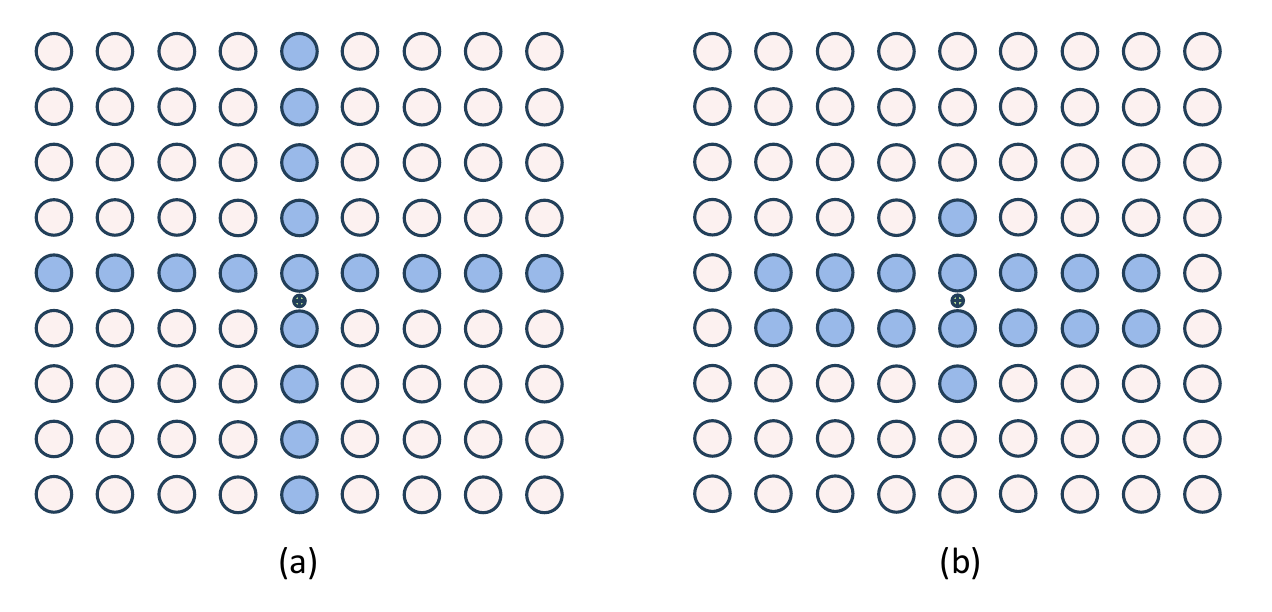} % \textwidth
	\caption{Illustration of the larger filter shape of CCALF with 25 taps.}
	\label{fig4}
\end{figure}
    
    {On the other hand, this contribution has also proposed to remove the power of 2 constraints on filter coefficient values, which was proposed again in the JVET meeting in \cite{ref17,ref18}.

    This contribution demonstrates significant gains in chroma components. However, the necessity of adaptively selecting CCALF shapes was questioned. In a subsequent exploration experience \cite{ref19}, in addition to the adaptive selection of the two filter shapes proposed in \cite{ref16}, another scheme involving larger size filters was proposed. Specifically, a 25-tap long-tap CCALF was introduced. This long-tap filter was considered a simpler scheme to achieve the gain. After joint tests of the modified CCALF and other in-loop filters \cite{ref20}, the long-tap CCALF scheme was eventually adopted. The new size of CCALF is illustrated in Fig.~\ref{fig5}.
    }
    
    {
    Because residual values have been stored and used in luma ALF, the concept of residual-based taps in chroma ALF and CCALF has also been proposed \cite{ref21}. Before this contribution, CCALF only had one online-trained CCALF filter with a cross-liked filter shape mentioned above, as depicted in Fig.~\ref{fig5}. Since the residual values are utilized in the unfixed luma filter of ALF, there is no need to store luma residual values additionally. In this contribution, only one luma-residual-based tap is added. Furthermore, chroma residual values are incorporated into the chroma online-trained filter of ALF, while luma residual values are employed in CCALF. However, considering that chroma residual values were not stored previously and the additional memory required, the resulting gain is comparatively low. Therefore, this proposal was recommended for further study.
    }
\begin{figure}[H]
	\centering
	\includegraphics[width=0.6\columnwidth]{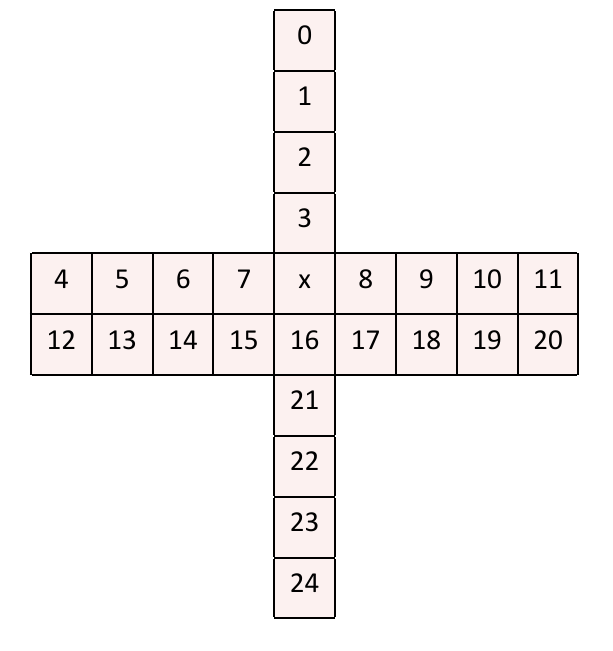} % \textwidth
	\caption{Illustration of the filter shape of CCALF with 25 taps.}
 %{Illustration of the filter shape of CCALF at the CTB level filter shape selection proposal.}
	\label{fig5}
\end{figure}
    {
    At the next round of the Meeting, the introduction of luma residual taps in chroma ALF and CCALF was proposed \cite{ref22}. Five luma residual taps in a cross 3x3 shape were added. These extended taps take the collocated and neighboring luma residual values as input. The inclusion of the luma residual taps in CCALF was adopted due to its relatively higher standalone gain \cite{ref23}. The filter shape of CCALF in ECM-12.0 is illustrated in Fig.~\ref{fig9}.

    The coefficients that need to be calculated are divided into two parts: spatial luma sample-based taps and luma residual-based taps. The linear filtering operation can be represented using Eqn.(\ref{eq1}).
    }

\begin{figure}[H]
	\centering
	\includegraphics[width=\columnwidth]{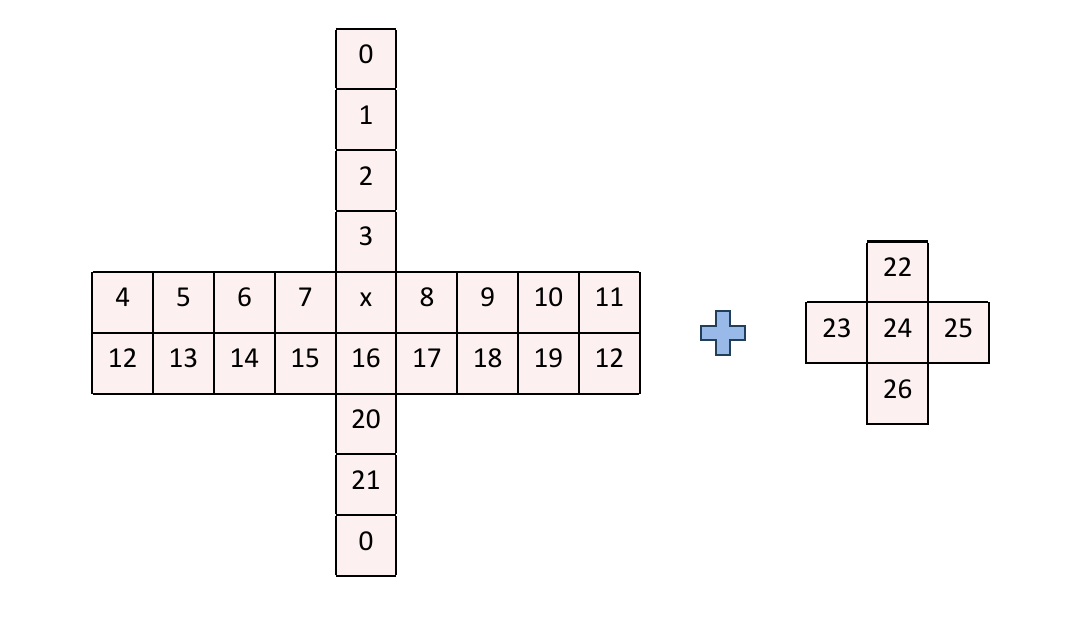} % \textwidth
	\caption{Illustration of the filter shape of CCALF at ECM-12.0. The left cross-liked filter uses the reconstructed spatial sample of luma SAO as input with 23 taps. The right one uses luma residual samples as input.
 }
	\label{fig9}
\end{figure}
    
    {At the 32nd JVET meeting, coefficient precision adjustment for ALF was proposed, demonstrating promising coding performance with negligible increases in encoding and decoding time \cite{ref25}. Similarly, at the 33rd JVET meeting, adaptive coefficient precision for CCALF was introduced \cite{ref17,ref18}. Since CCALF involves different coefficient derivation compared to ALF, removing the power of 2 constraints was also proposed in this context. This adjustment can enhance the accuracy of coefficients, though a two-bit syntax element needs to be signaled for each set of luma filter sets to indicate the number of bits. These two contributions have been further investigated.
    }
\fi

\subsection{Filter Shape}

{The filter shape of CCALF was a $5\times6$ diamond-shaped filter with 14 filter coefficients and 18 taps when it was initially proposed~\cite{o0636}. Considering the trade-off among performance, line buffer, and computational complexity, several reduced filter shapes were proposed~\cite{p0558, p0106, p0173, p0251}. Finally, the $3\times4$ diamond filter shape was adopted in H.266/VVC. Fig.~\ref{fig3} illustrates the relative location of the chroma sample being filtered and its support region in the luma sample when CCALF is adopted in H.266/VVC. Consequently, each CCALF filter has only 7 filter coefficients, the filtering operation is shown in Eqn.(\ref{eq1}) and $N=7$. %In ALF, the number of coefficients is reduced by utilizing symmetry. However, CCALF does not adopt this method due to its potential impact on compression efficiency. Instead, CCALF employs a more restricted symmetry scheme at the tips of the diamond.
}

\begin{figure}[H]
	\centering
	\includegraphics[width=\columnwidth]{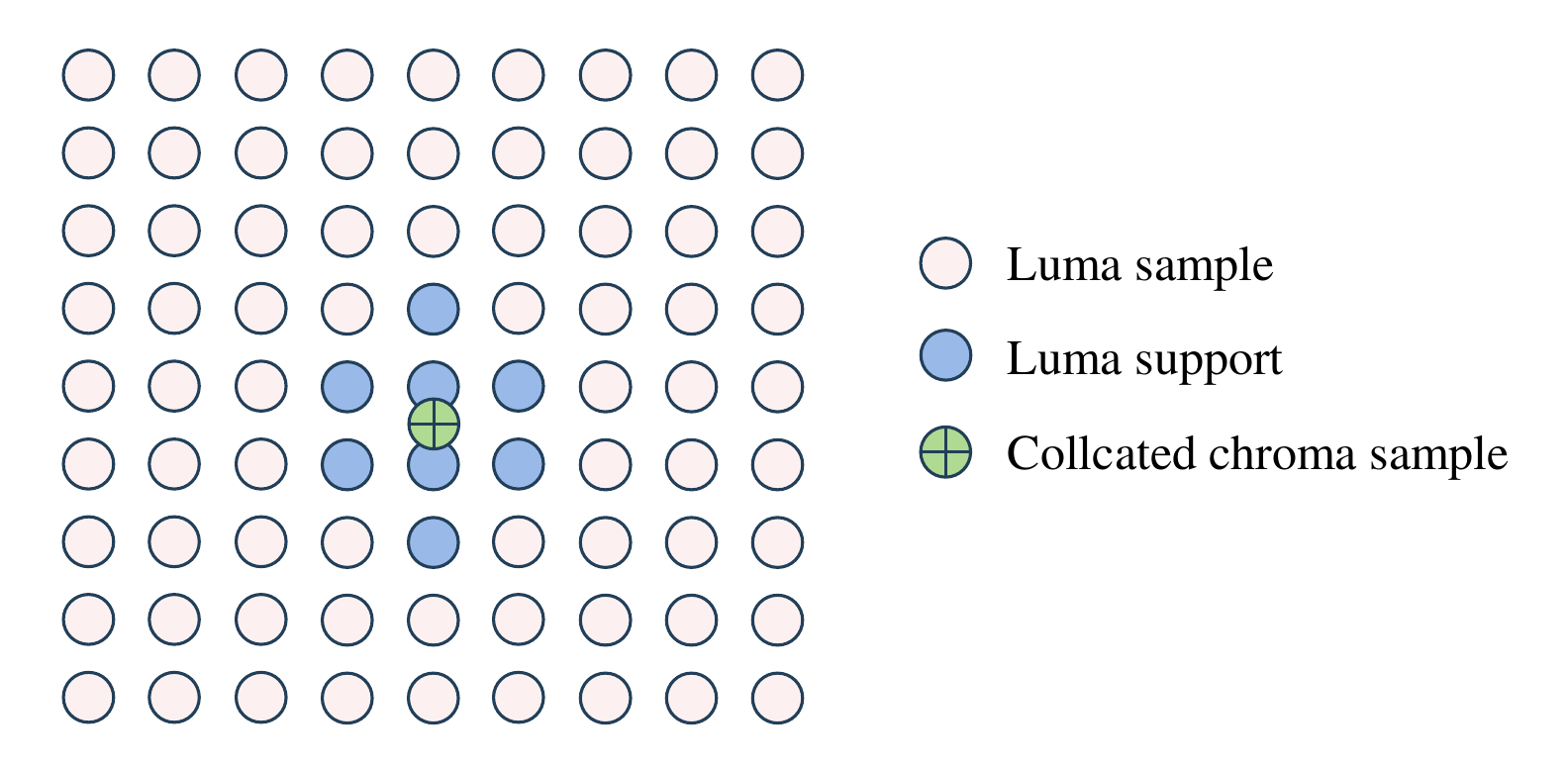} % \textwidth
	\caption{Illustration of the relative location of filtered chroma sample of CCALF and its  support in the luma channel for 4:2:0  chroma format in H.266/VVC.}
	\label{fig3}
\end{figure}

{To further improve the performance of CCALF, numerous proposals related to CCALF have been put forward, considering the trade-off between performance and running time, complexity, or other factors. However, some of these proposals were not adopted.

In \cite{V0080}, an extension to CCALF has been proposed. This contribution suggests extensions to CCALF in both the number and size of filters. While this extension can enhance chroma components, it may lead to some loss in the luma component. Additionally, CCALF could introduce artifacts in chroma components, which is why certain constraints are set in high QP regions. Therefore, proposals regarding CCALF must avoid reintroducing these artifacts.

Considering that the correlation between neighboring pixels may depend on the characteristics of the video content, a single filter shape may not be optimal for different video content. To further enhance the performance of CCALF, a Coding Tree Block~(CTB) level filter shape selection scheme was proposed~\cite{ref16}. This contribution introduces two filter shapes, as shown in Fig.~\ref{fig4}. Within each Adaptation Parameter Set~(APS), multiple filters and their corresponding shapes with coefficients are signaled. For each CTB, the decoder specifies which filter shape and coefficients are used based on the signaled index. 
}

\begin{figure}[H]
	\centering
	\includegraphics[width=\columnwidth]{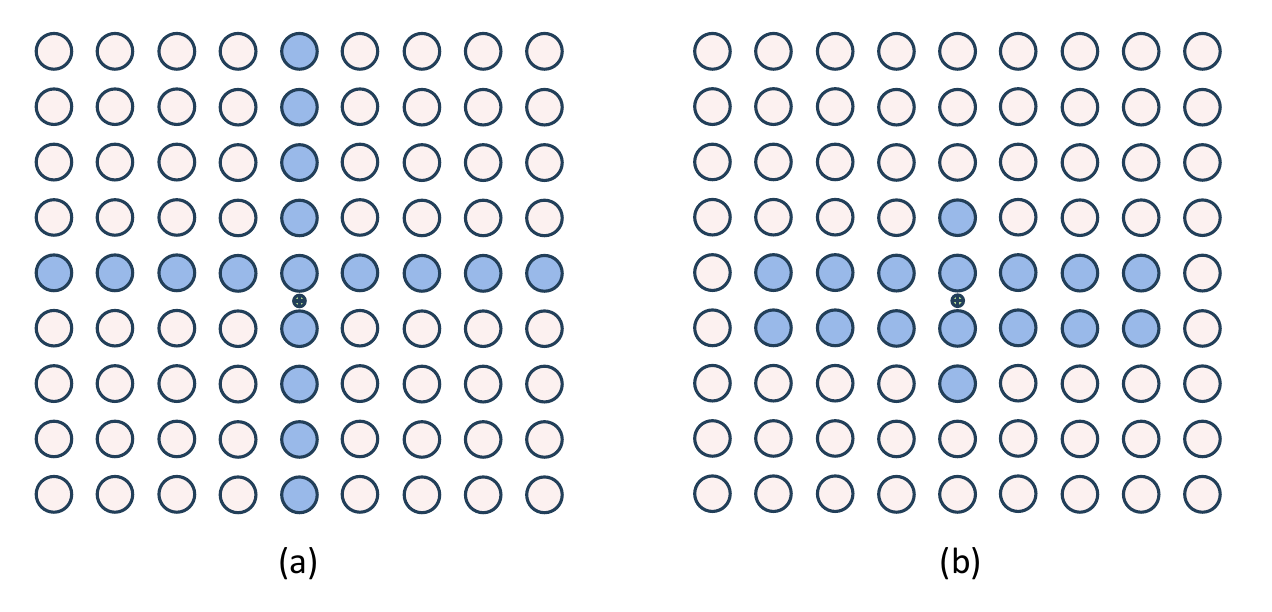} % \textwidth
	\caption{Illustration of the larger filter shape of CCALF with 25 taps.}
	\label{fig4}
\end{figure}

This contribution demonstrates significant gains in chroma components. However, the necessity of adaptively selecting CCALF shapes was questioned. In a subsequent exploration experience \cite{ref19}, in addition to the adaptive selection of the two filter shapes proposed in \cite{ref16}, another scheme involving larger-size filters was proposed. Specifically, a 25-tap long-tap CCALF was introduced. This long-tap filter was considered a simpler scheme to achieve better gain. After joint tests of the modified CCALF and other in-loop filters \cite{ref20}, the long-tap CCALF scheme was eventually adopted. The new shape of CCALF in ECM is illustrated in Fig.~\ref{fig5}, the filtering operation is shown in Eqn.(\ref{eq1}) and $N=25$.

{
Because residual values have been stored and used in luma ALF, the concept of residual-based taps in chroma ALF and CCALF has also been proposed \cite{ref21}. Before this contribution, CCALF only had one online-trained CCALF filter with a cross-liked filter shape mentioned above, as depicted in Fig.~\ref{fig5}. Since the residual values are utilized in the unfixed luma filter of ALF, there is no need to store luma residual values additionally. In this contribution, only one luma-residual-based tap is added. Furthermore, chroma residual values are incorporated into the chroma online-trained filter of ALF, while luma residual values are employed in CCALF. However, considering that chroma residual values were not stored previously and the additional memory required, the resulting gain is comparatively low. Therefore, this proposal was recommended for further study.
}

\begin{figure}[H]
	\centering
	\includegraphics[width=0.6\columnwidth]{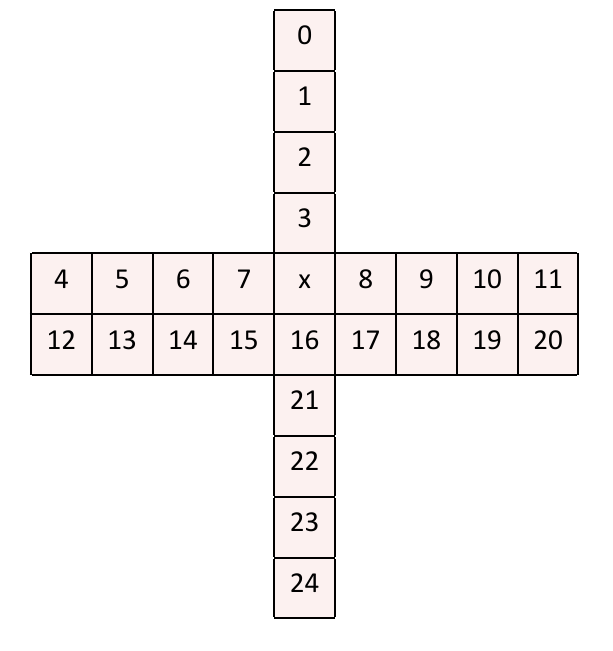} % \textwidth
	\caption{Illustration of the filter shape of CCALF with 25 taps.}
 %{Illustration of the filter shape of CCALF at the CTB level filter shape selection proposal.}
	\label{fig5}
\end{figure}
    {
    At the 31st JVET meeting, the introduction of luma residual taps in chroma ALF and CCALF was proposed \cite{ref22}. Five luma residual taps in a cross 3x3 shape were added. These extended taps take the collocated and neighboring luma residual values as input. The inclusion of the luma residual taps in CCALF was adopted due to its relatively higher standalone gain \cite{ref23}. The filter shape of CCALF in ECM-12.0 is illustrated in Fig.~\ref{fig9}.

The coefficients that need to be calculated are divided into two parts: spatial luma sample-based taps and luma residual-based taps. The linear filtering operation can be represented using Eqn.(\ref{eq2}).
}
{ \scriptsize

    \begin{equation}    
        \begin{aligned}
           {\tilde{I}(x, y) = I(x, y)+\sum_{i = 0, 12} c_{i}\left(f_{i, 0}+f_{i, 1}\right)}\\
           {+\sum_{i = 1}^{21} c_{i} f_{i}+\sum_{i = 22}^{26} c_{i} g_{i},}
        \end{aligned}
            %{\tilde{I}(x, y) = I(x, y)+\sum_{i = 0}^{n-1} c_{i} f_{i}}
        \label{eq2}
    \end{equation} 
    \begin{equation}    
        \begin{aligned}
           {f_{i, j}=L(x'+x_{i,j}, y'+y_{i,j})-I(x, y),}
        \end{aligned}
        %\label{eq2}
    \end{equation}
    \begin{equation}    
        \begin{aligned}
           {g_{i}=Clip(R(x'+x_{i}, y'+y_{i})),}
        \end{aligned}
        %\label{eq2}
    \end{equation}
}
{\noindent
where $(x,y)$ is the coordinate of the center sample, $(x',y')$ is the coordinate of collocated luma sample. $(x'+x_{i}, y'+y_{i})$ and $(x'+x_{i,j}, y'+y_{i,j})$are the coordinates of the reconstructed samples corresponding to coefficient $c_{i}$, $f_{i,j}$ and $f_{i}$ is the difference between neighboring luma samples $L(x',y')$ and current sample $I(x, y)$, $g_{i}$ is the clipped value of luma residual samples $R(x',y')$ which is the residual between prediction samples and reconstructed samples. $Clip$ is the function that limits the values within a certain range to reduce the impact of significant differences in sample values, the value of the clipping operation depends on the clipIdx of APS and bitdepth.
}

\begin{figure}[H]
	\centering
	\includegraphics[width=\columnwidth]{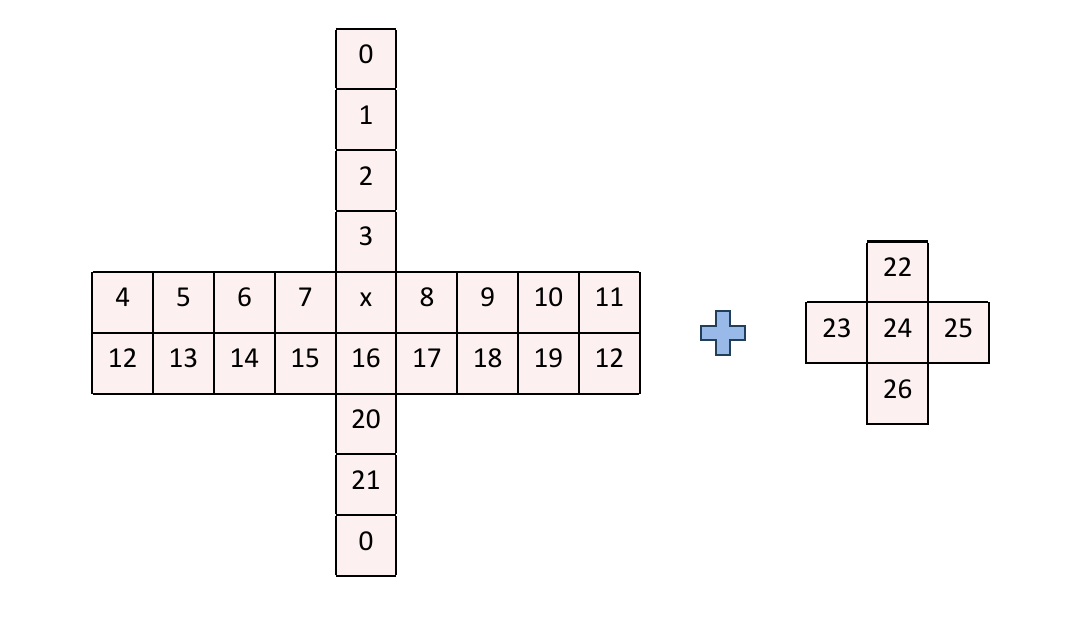} % \textwidth
	\caption{Illustration of the filter shape of CCALF at ECM-12.0. The left cross-liked filter uses the reconstructed spatial sample of luma SAO as input with 23 taps. The right one uses luma residual samples as input.
 }
	\label{fig9}
\end{figure}

\subsection{Filter Coefficient Calculation and Representation}

Except for the filter shape of the CCALF, the optimization of coefficient calculation and signaling~\cite{o0636,p0557,q0165,r0322,r0327} is also an important topic to improve the performance of CCALF.

When CCALF was proposed, each filter had 14 filter coefficients and 18 taps, and every coefficient has an 8-bit dynamic range and is signaled with a third-order exponential-Golomb code~\cite{o0636}. However, it would increase complexity with additional multiplications per chroma pixels. To simplify the computation overhead, a bit shifting scheme was proposed to replace the multiplications~\cite{p0557}.  The results show that this scheme can reduce the complexity of the CCALF filter with accepted loss,  so it was adopted. Besides, a contribution was proposed to reduce memory access, encoding latency, and power consumption~\cite{r0327}.  It proposes a method to estimate CCALF filtering distortion without conducting real filter operations. With this proposal, the number of encoding passes can be reduced from 152 to 1 while did not affect the coding performance. As a desirable simplification, this proposal was adopted. 

%There are also some adopted contributions designed to fix the bug in the coefficient calculation of CCALF~\cite{q0165,r0322}.

%On the other hand, this contribution has also proposed to remove the power of 2 constraints on filter coefficient values, which was proposed again in the JVET meeting in \cite{ref17,ref18}.

{At the 32nd JVET meeting, coefficient precision adjustment for ALF was proposed, demonstrating promising coding performance with negligible increases in encoding and decoding time \cite{ref25}. Similarly, at the 33rd JVET meeting, adaptive coefficient precision for CCALF was introduced \cite{ref17,ref18}. Since CCALF involves different coefficient derivation compared to ALF, removing the power of 2 constraints was also proposed in this context. This adjustment can enhance the accuracy of coefficients, though a two-bit syntax element needs to be signaled for each set of luma filter sets to indicate the number of bits. These two contributions have been further investigated.}

\subsection{Syntax Design}
    {
    %In H.266/VVC, CCALF was introduced as an in-loop filter, utilizing the reconstructed luma sample of SAO as input for an adaptive linear filter to refine the reconstructed chroma sample of Chroma-ALF. CCALF in H.266/VVC uses the 3$\times$4 diamond filter, as shown in Fig.~\ref{fig3}. Unlike in H.266/VVC, ECM12.0 utilizes luma residual samples additionally, as shown in Fig.~\ref{fig9}. The residual correction is generated for chroma samples according to Eqn.(\ref{eq1}).
    Compared to H.266/VVC, ECM-12.0 utilizes luma residual samples additionally, as shown in Fig.~\ref{fig9}. The residual correction is generated for chroma samples according to Eqn.(\ref{eq2}).
    %For each picture, CCALF can transmit up to 8 CCALF filters, with the resulting filters then indicated for each of the two chroma channels on a CTU basis. Each slice only has one APS, and the Cb component and Cr component can have different APSs, which are signaled separately at the slice header. Similar to luma ALF, to reduce bit overhead, filter coefficients of different classifications can be merged.
    %check:
    For each picture, two types of information need to be coded for CCALF, i.e., filter coefficient parameters and filter control on/off flags. The filter coefficient parameters include the number of cross-component filters and the coefficients of the corresponding filter. CCALF can transmit up to 8 CCALF filters, with the resulting filters then indicated for each of the two chroma channels on a CTU basis. Each slice only has one APS, and the Cb component and Cr component can have different APSs, which are signaled separately at the slice header. Similar to luma ALF, to reduce bit overhead, filter coefficients of different classifications can be merged. The filter control on/off flags are used to provide better local adaptation, there are sequence-level, picture-level, slice-level and CTU-level filter on/off control. When the value of sequence-level and picture-level control flags is not present, it is inferred to be equal to 0. When slice-level on/off control flag is not present, it is inferred to be equal to picture-level on/off control flags. If the slice-level on/off control flag indicates ALF-on, CTU-level filter on/off control flags are interleaved in slice data and coded with CTUs; otherwise, no additional CTU-level filter on/off control flags are coded and all CTUs of the slice are inferred as ALF-off.
    
    %Considering that CCALF may introduce artifacts in high QP regions, to avoid a reduction in subjective quality, CCALF will be disabled when any of the following conditions are true:
    Due to the abundant texture features of the luma component, CCALF may introduce artifacts with overly abundant chroma texture, thereby reducing the subjective quality of the image, especially at high QP. Therefore, the H.266/VVC reference encoder can achieve subjective tuning by configuring the config file. Specifically, it can attenuate the application of CCALF in high QP encoding and areas with high-frequency luminance. Algorithmically, CCALF is deactivated on CTUs when any of the following conditions are true,
    
    (1) The slice QP value minus 1 is less than or equal to the base QP value.
    
    (2) The number of chroma samples for which the local contrast is greater than $( 1 << ( bitDepth-2 ) )-1 $ exceeds the CTU height, where the local contrast is the difference between the maximum and minimum luma sample values within the filter support region.
    
    (3) More than a quarter of chroma samples are in the range between $( 1 << ( bitDepth -1 ) )-16$ and $( 1 << ( bitDepth -1 ) ) + 16$.
 
    }
   
\section{CCSAO}
    {CCSAO is conceptually similar to SAO, as it initially classifies the samples to be filtered into different categories, then derives an offset value for each category, and finally corrects the pixels in that category with the corresponding offset value. It uses the reconstructed sample of DBF, which is the same as SAO, the offsets are derived for three channels respectively. The reconstruction operation of CCSAO can be represented by the equation below,
{ \scriptsize
\begin{equation}    
    %\begin{aligned}
        {\tilde{C_{rec}} = Clip(C_{rec}+offset_{i}),}
    %\end{aligned}
    \label{eq8}
\end{equation}   
}%
    $C_{rec}$ and $\tilde{C_{rec}}$ are the reconstructed sample after DBF and CCSAO, respectively. $i$ represents the class index of the corresponding sample, $offset_{i}$ is the corresponding offset value.
    
    %is the reconstructed sample after DBF, The $\tilde{C_{rec}}$ is the reconstructed sample after CCSAO, the $i$ is the class index of this sample, the $offset_{i}$ is the corresponding offset value.

    The difference between SAO and CCSAO lies in CCSAO's utilization of the strong correlation between the luma and chroma components in the classification process. It optimizes the reconstruction of one component of the sample by leveraging the information contained in the other component of the sample \cite{ref9}.
}

\subsection{Classifier Extension}
    {    
    The original CCSAO includes only a classification based on band information to avoid a significant increase in complexity. Corresponding band offsets are obtained by minimizing the sum of squared error (SSE) between the original sample and the corrected reconstruction sample. This approach keeps computational complexity low while enabling CCSAO to handle certain encoded artifacts.
    %that ALF and CCALF cannot, such as mean deviations. 
    It should be noticed that the offsets need to be signaled in the bitstream.

    CCSAO is applied to the output of DBF reconstructed samples, and the offset calculated for each category is added to the output sample from the SAO process. Therefore, CCSAO can be parallelized with SAO, as shown in Fig.~\ref{fig6}. 
\begin{figure}[H]
	\centering
	\includegraphics[width=\columnwidth]{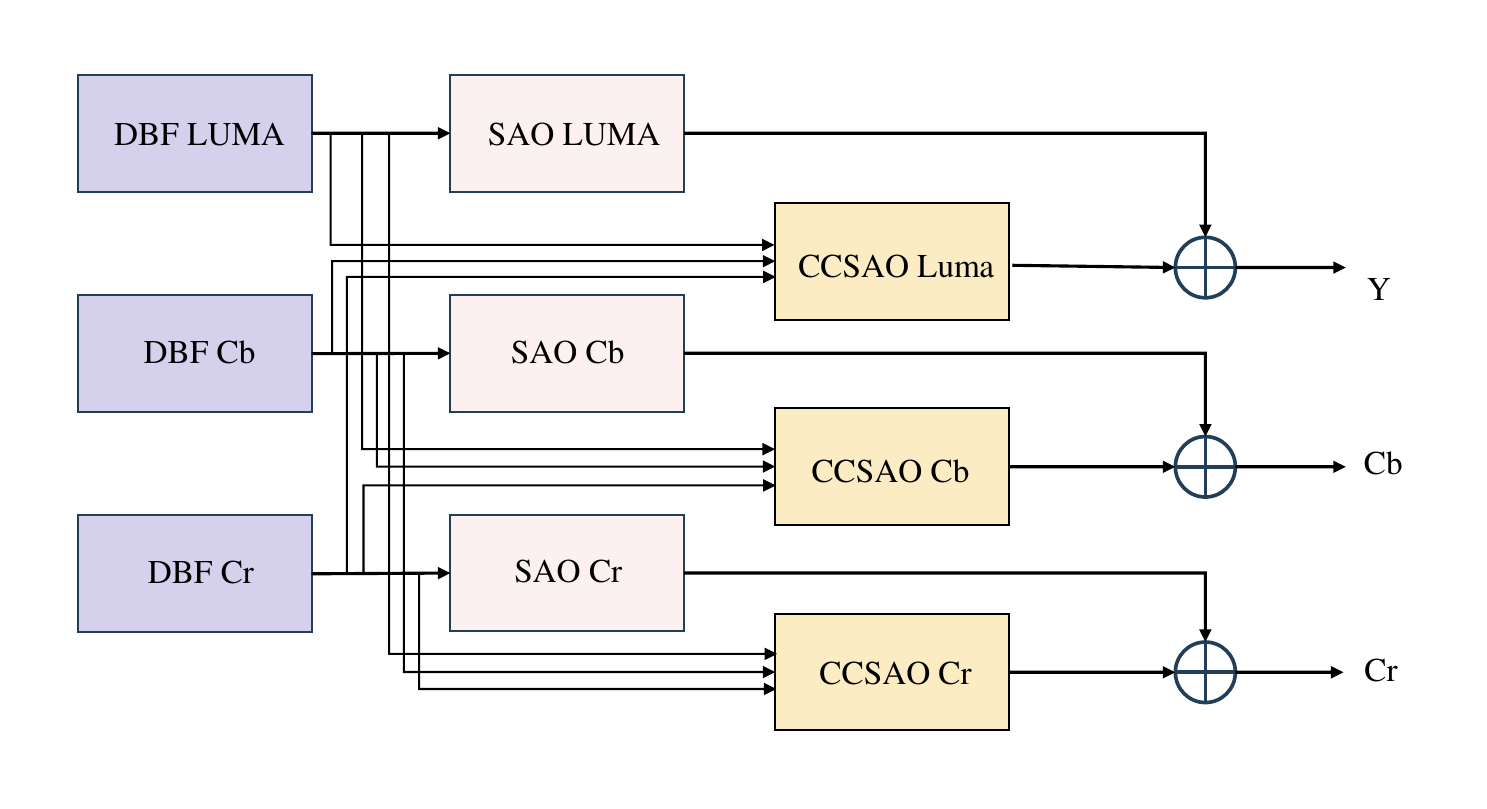} % \textwidth
	\caption{Illustration of SAO process when CCSAO is applied.}
	\label{fig6}
\end{figure}

    The band information-based classification of CCSAO utilizes the reconstructed sample of three components to process the classification for each component. Specifically, the collocated samples for each component are first selected. Then, an index representing a category is calculated based on the band number of the three components and their collocated samples. The offset value of a sample depends on its category. Regarding the collocated samples for each component, the collocated luma sample can be chosen from 9 candidates, while the collocated chroma samples have fixed positions, as shown in Fig.~\ref{fig7}.
\begin{figure}[H]
	\centering
	\includegraphics[width=\columnwidth]{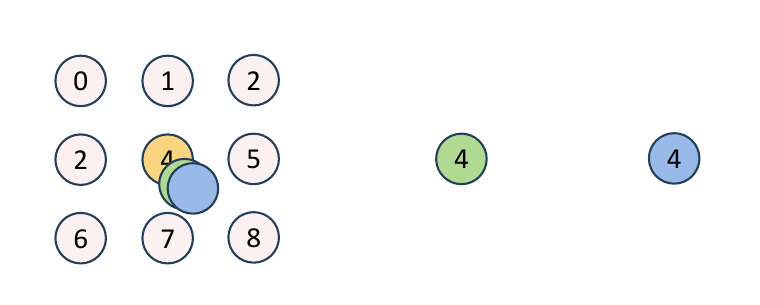} % \textwidth
	\caption{Illustration of the collocated sample used for the CCSAO classification. The left graph shows the 9 locations of the luma component, one of the 9 samples will be chosen based on RDO, and the green and blue samples show the two collocated chroma samples respectively.}
	\label{fig7}
\end{figure}    
    
    CCSAO was first proposed and adopted\cite{ref12} in the AVS3 video coding standard, in which collocated luma component samples are classified by equally dividing the range of the sample values. For each category, an offset value is derived and used for the chroma samples whose collocated luma sample belongs to the category.

    Although cross-component tools in in-loop filters always act on chroma components, regarding cross-component proposals, attention should not only be given to the gain of chroma components but also to the effects on the luma component. Furthermore, subjective quality improvement needs to be considered as well.
    Considering these reasons, CCSAO was introduced to ECM. This proposal showed great performance improvement in chroma components while introducing minimal loss in the luma component. Initially, CCSAO only used the band classifier when it was adopted in ECM \cite{ref26},
    %update：
     and the category index will be calculated using the equation below,\par
    
    { \scriptsize
    \begin{equation}    
        %\begin{aligned} 
            {classIndex=band_{Y} * (N_{Cb}*N_{Cr}) + band_{Cb} * N_{Cr} + band_{Cr}},
        %\end{aligned}
    \end{equation}
    \begin{equation}    
        %\begin{aligned} 
            {band_{L}=P(x_{Y}, y_{Y}) * N_{Y} >> BitDepth},
        %\end{aligned}
    \end{equation}
    \begin{equation}    
        %\begin{aligned} 
            {band_{Cb}=P(x_{Cb}, y_{Cb}) * N_{Cb}>> BitDepth},
        %\end{aligned}
    \end{equation}
    \begin{equation}    
        %\begin{aligned} 
            {band_{Cr}=P(x_{Cr}, y_{Cr}) * N_{Cr}>> BitDepth},
        %\end{aligned}
    \end{equation}
    
    }%

    {\noindent
    where 
    $P(i, j)$ is the sample value of different component at position $(i, j)$, $N_{i}$ is the number of band for each component, $(x_{C}, y_{C})$ is the current chroma sample position, $x_{L}, y_{L}$ is the collocated luma sample position.
    }
%update end    
    As a new in-loop filter tool, several schemes have been proposed to optimize the original CCSAO. An extension of CCSAO was proposed at the 24th of JVET meetings, where the proponents extended the design of CCSAO by adding the edge-based classifier \cite{X0105,X0152}. Similar to the edge-based classification method in SAO, the edge-based classification of CCSAO also uses four 1-D directional patterns, including horizontal, vertical, 45°, and 135°, as shown in Fig.~\ref{fig8}. The best direction mode is determined at the encoder through rate-distortion optimization (RDO). Edge information used for classification is derived by calculating the difference between the center pixel and its two adjacent pixels, and then comparing the difference with a predefined threshold value to derive the final class index. The best threshold values are also selected from an array of predefined threshold values based on RDO.
%update:
    If the edge-based classifier is selected, the category index will be calculated as follows, given the chroma sample and the collocated luma samples,\par
     { \scriptsize
   
    \begin{equation}    
        \begin{aligned}
            {classIndex = BandNum \ast  16 + q_{a} \ast  4 + q_{b}},
        \end{aligned}
        \label{eq8}
    \end{equation}
    \begin{equation}    
        %\begin{aligned}
            %{q_{a}=(d_{a}<0)? (d_{a}<-Th ? 0:1):(d_{a}<Th ? 2:3) },
            {q_{i} =\left\{ 
                \begin{array}{lc}
                    0 & d_{i}<-Th, \\
                    1 & -Th<d_{i}<0,\\
                    2 & 0<d_{i}<Th,\\
                    3 & Th<d_{i}.\\
                \end{array}
            \right.}
        %\end{aligned}
    \end{equation}
    %\begin{equation}    
        %\begin{aligned}
            %{q_{b}=(d_{b}<0)? (d_{b}<-Th ? 0:1):(d_{b}<Th ? 2:3)},
    %       {q_{b} =\left\{ 
    %          \begin{array}{lc}
    %               0 & d_{b}<-Th \\
    %                1 & -Th<d_{b}<0\\
    %                2 & 0<d_{b}<Th\\
    %                3 & Th<d_{b}\\
    %            \end{array}
    %        \right.},
        %\end{aligned}
    %\end{equation}  
    }\par
    
    { \scriptsize
    \begin{equation}    
        %\begin{aligned}
            %{q_{a}=(d_{a}<0)? (d_{a}<-Th ? 0:1):(d_{a}<Th ? 2:3) },
            {BandNum = cur_{i} * N_{i} >> BitDepth},
        %\end{aligned}
        \label{eq9}
    \end{equation} 
    }% 
    {\noindent
    where 
    $i$ can be chosen from the two co-located samples based on RDO,
    $d_{i}$ is the delta value between the center sample $c$ and the neighboring sample $a$ or $b$. $q_{i}$ is the quantized value of $d_{i}$. %Variable $BandNum$ in Eqn.(\ref{eq8}) is derived as follow,
    }
    The position of neighboring sample $a$ or $b$ depends on the best 1-D directional pattern selected from the four 1-D directional patterns. Besides, the range of the offset value is also constrained to $[-15, 15]$, and these offsets are needed to transmit to the decoder.
%update end    
    
    Unlike SAO, the edge-based classifier in CCSAO combines luma edge and the band index of the sample at the corresponding collocated position to determine the final classification of a given sample. Additionally, CCSAO uses collocated luma samples to derive edge information for chroma samples, while SAO uses neighboring samples of the same component to derive edge information.
    
    A similar contribution has been proposed at AVS \cite{ref15}, the Enhanced Cross-Component Sample Adaptive Offset (ECCSAO) method further improves encoding performance, which includes an extension of the edge-based classification method, using the edge information of collocated luma samples to classify chroma samples. Moreover, a four-layer quad-tree structure is proposed. The former method has been adopted by AVS.
    }
    
\begin{figure}[H]
	\centering
	\includegraphics[width=\columnwidth]{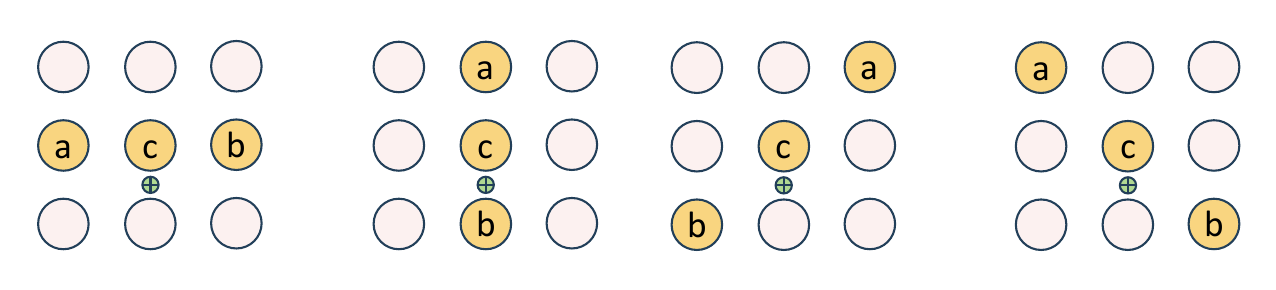} % \textwidth
	\caption{Illustration of the edge-based classification of CCSAO. Four graphs show four different directions, the yellow samples are the locations used for calculating the class index at different directional patterns.}
	\label{fig8}
\end{figure}
     
    {
    In the ECM, the edge classifier was further optimized with more edge/band combinations, and the component used for edge classification can be selected from any of the three components \cite{ref11,ref24}. The new edge-based classification scheme, which is a subset of the original one with fewer edge range divisions was added. This allows for more flexible edge/band combinations to adapt to the local characteristics of video sequences. This contribution was adopted at the 31st JVET meeting.}
%update
    The second edge-based classifier is formulated as follows,
    { \scriptsize
    \begin{equation}    
        \begin{aligned}
            {classIndex = BandNum \ast  4 + q_{a} \ast  2 + q_{b}},
        \end{aligned}
    \end{equation} 
    \begin{equation}    
        %\begin{aligned}
            %{q_{a}=(d_{a}<0)? (d_{a}<-Th ? 0:1):(d_{a}<Th ? 2:3) },
            {q_{i} =\left\{ 
                \begin{array}{lc}
                    0 & d_{i}<-Th,\\
                    1 & -Th<d_{i}<0.\\
                \end{array}
            \right.}
        %\end{aligned}
    \end{equation}      
    }%
%update end
\subsection{Signaling Overhead Reduction}    
{ 
    {Similar to the APS design in H.266/VVC, the inheritance scheme of CCSAO was also proposed in \cite{ref11,ref24}. It was noted that there is a strong correlation between the CCSAO offsets and classifier parameters of different pictures. To reduce signaling overhead, the offsets/parameters of some coded pictures can be stored at both the encoder and decoder, allowing them to be used by future pictures. This contribution has also been adopted.
    }
    
    %{In \cite{ref13}, a cross-component in-loop tool named Cross-Component Sample Offset (CCSO) is proposed. CCSO is a non-linear loop filtering approach. The filtering of CCSO utilizes a diamond-shaped filtering support that includes 5 samples as input. The collocated reconstructed luma sample of the current chroma sample to be filtered, and the four surrounding samples of the collocated reconstructed luma are used as input. The delta value between the four surrounding samples and the center luma sample is further quantized. The offsets are derived through a mapping process using a look-up table with the quantized delta value as input.}
}

\if false    
\subsection{CCSAO in ECM}
    During the development of ECM, CCSAO has been adopted due to its excellent coding performance.
    Two modes can be selected in ECM: band mode, which is based on texture information, and edge mode, which is based on edge information.

    If the edge mode is selected, the category index will be calculated as follows, given the chroma sample and the collocated luma samples,\par
     { \scriptsize
   
    \begin{equation}    
        \begin{aligned}
            {classIndex = BandNum \ast  16 + q_{a} \ast  4 + q_{b}},
        \end{aligned}
        \label{eq8}
    \end{equation}
    \begin{equation}    
        %\begin{aligned}
            %{q_{a}=(d_{a}<0)? (d_{a}<-Th ? 0:1):(d_{a}<Th ? 2:3) },
            {q_{i} =\left\{ 
                \begin{array}{lc}
                    0 & d_{i}<-Th, \\
                    1 & -Th<d_{i}<0,\\
                    2 & 0<d_{i}<Th,\\
                    3 & Th<d_{i}.\\
                \end{array}
            \right.}
        %\end{aligned}
    \end{equation}
    %\begin{equation}    
        %\begin{aligned}
            %{q_{b}=(d_{b}<0)? (d_{b}<-Th ? 0:1):(d_{b}<Th ? 2:3)},
    %       {q_{b} =\left\{ 
    %          \begin{array}{lc}
    %               0 & d_{b}<-Th \\
    %                1 & -Th<d_{b}<0\\
    %                2 & 0<d_{b}<Th\\
    %                3 & Th<d_{b}\\
    %            \end{array}
    %        \right.},
        %\end{aligned}
    %\end{equation}  
    }%
    {\noindent
    where 
    $d_{i}$ is the delta value between the center sample $c$ and the neighboring sample $a$ or $b$. $q_{i}$ is the quantized value of $d_{i}$. Variable $BandNum$ in Eqn.(\ref{eq8}) is derived as follow,
    }
    { \scriptsize
    \begin{equation}    
        %\begin{aligned}
            %{q_{a}=(d_{a}<0)? (d_{a}<-Th ? 0:1):(d_{a}<Th ? 2:3) },
            {BandNum = cur_{i} * N_{i} >> BitDepth},
        %\end{aligned}
        \label{eq9}
    \end{equation} 
    }% 
    {\noindent
    where 
    $i$ can be chosen from the current sample being processed and the two co-located samples based on RDO.
    }
\fi    
    
    %{\noindent
    %The position of neighboring sample $a$ or $b$ depends on the best 1-D directional pattern selected from the four 1-D directional patterns by RDO. Besides, the range of the offset value is also constrained to $[-15, 15]$, and these offsets are needed to transmit to the decoder.
    %}
 
\section{Performance Evaluation}

    To improve the coding performance, both CCALF and CCSAO are integrated into ECM-12.0 seamlessly. 
    A comparative analysis is conducted to evaluate the efficiency and effectiveness of the cross-component in-loop filter tools.
    With continuous development, both CCSAO and CCALF have achieved remarkable performance gains.
    To evaluate the coding performance of CCALF and CCSAO, ECM-12.0 without CCALF and CCSAO are regarded as anchor respectively~\cite{ctc}.
    
    As shown in Table~\ref{table:1}, CCALF can achieve 2.49\% and 2.90\% coding gains for Cb and Cr components under AI configuration.
    For RA configuration, 1.48\% and 2.12\% coding gains for Cb and Cr components can be achieved.
    In VTM-10.0, CCALF can achieve 13.88\% and 13.73\% coding gains for Cb and Cr components under AI configuration, and 9.69\% and 8.55\% coding gains for Cb and Cr components under RA configuration. The decrease in gain may caused by the newly proposed and optimized cross-component techniques in the prediction process.
    %One can also discern that Y component also achieves xx\% coding performance gain, which can be attribute to the bitrate reduction of the chroma component.
    For CCSAO, 1.28\% and 1.08\% coding gains can be achieved for Cb and Cr components under AI configuration.
    For RA configuration, 3.02\% and 2.79\% coding gains for Cb and Cr components can be achieved.
    
    It can be noted that the Y component coding performance of CCSAO on screen content sequences is significantly greater than that of natural sequences. This may be caused by the more obvious relationship between the texture and directional features of luma and chroma components in screen content videos.
    Furthermore, we also compared the subjective performance under different configurations. The subjective testing materials consist of the sequences mentioned in CTC and each sequence has been encoded with four QPs(QP=22, 27, 32 and 37) under AI and RA configurations. Partial visual quality comparison results of reconstructed sequences are shown in Fig.~\ref{}. In Fig.~\ref{}, the decoded images with CCALF and CCSAO are provided in the first column, the decoded images without CCALF, the last column shows the reconstructed image without CCSAO. The red boxes are highlighted to indicate the region with significant subjective improvement.

    %To demonstrate the capability of cross-component tools more intuitively, we compare the performance between the ECM12.0 reference software with and without the cross-component in-loop filter. 
    
    %The anchor we chose is ECM12.0, and the tests involve ECM12.0 with CCSAO turned off and ECM12.0 with CCALF turned off, respectively.
    
    %On average, CCALF can achieve a performance improvement which has 2.49\% on Cb component and 2.90\% on Cr component at all intra configuration, 
    %CCSAO can achieve a performance improvement which has 1.28\% on Cb component and 1.08\% on Cr component at all intra configuration. 
\subsection{the trend of loop filters}
{
    The loop filters are designed to correct artifacts that are introduced before loop filtering. Different kinds of loop filters can deal with different artifacts like blocking, ringing, blurring, mosquito, etc. There are three kinds of crucial loop filters in VVC in total, namely DBF, SAO, and ALF. In VVC, the cross-component loop filter called CCALF is introduced to ALF to utilize the relationship between luma and chroma components fully. To further exploit the relationship, another cross-component loop filter is proposed for ECM named CCSAO, which is paralleled with SAO. With the development of ECM, the classifiers of CCSAO become more refined and diverse~\cite{X0105,X0152,ref11,ref24,ref10}, the structure of CCALF become more complex and comprehensive, more kinds of samples were added to the filters of CCALF~\cite{ref21,ref22}, the shape and calculation method of filters are also constantly being optimized~\cite{p0558,p0106,p0173,p0251,p0557,ref17}.
    Besides, some other in-loop filters based on the image non-local similarity have been studied~\cite{nonlocal3,nonlocal2,nonlocal1,meng2022parametric,non-local3} because the loop filter in existing video coding standards focus only on the local correlation.
    
    While these non-local loop filters can yield certain performance gains, their high computational demands and hardware limitations render their application in video coding standards challenging. Therefore, relevant methods to optimize non-local filters are exploring~\cite{non-opt,jia2020fast}
}

% Table generated by Excel2LaTeX from sheet 'Sheet2'
\begin{table*}[!t]
  \centering
  %\caption{ECM12.0 without CCALF vs. ECM12.0}
  \caption{Experimental Results of ECM-12.0, Anchor: ECM-12.0 Without CCALF}
  \resizebox{0.9\linewidth}{!} {
    \begin{tabular}{cccccc|ccccc}
    \toprule
    \multirow{2}[4]{*}{Class} & \multicolumn{5}{c|}{AI}               & \multicolumn{5}{c}{RA} \\
\cmidrule{2-11}          & Y     & Cb     & Cr     & EncT  & DecT  & Y     & Cb     & Cr     & EncT  & DecT \\
    \midrule
    A1    & 0.09\% & -1.21\% & -3.32\% & 99.8\% & 102.3\% &       &       &       &       &  \\
    A2    & 0.11\% & -2.78\% & -3.23\% & 94.2\% & 91.7\% &       &       &       &       &  \\
    B     & 0.12\% & -3.35\% & -3.22\% & 98.7\% & 97.6\% &       &       &       &       &  \\
    C     & 0.10\% & -1.67\% & -1.91\% & 99.4\% & 99.6\% & 0.03\% & -1.48\% & -2.12\% & 98.7\% & 97.8\% \\
    E     & 0.15\% & -3.12\% & -2.96\% & 96.1\% & 96.2\% &       &       &       &       &  \\
    \midrule
    Average & 0.11\% & -2.49\% & -2.90\% & 97.8\% & 97.6\% &       &       &       &       &  \\
    \midrule
    D     & 0.02\% & -0.42\% & -0.18\% & 96.6\% & 92.2\% & -0.01\% & -0.94\% & -0.53\% & 105.0\% & 105.5\% \\
    F     & 0.10\% & -1.77\% & -1.07\% & 97.0\% & 98.5\% & 0.15\% & -1.08\% & -0.32\% & 99.6\% & 100.3\% \\
    TGM   & 0.12\% & -1.19\% & -0.72\% & 96.5\% & 97.2\% & 0.16\% & -1.26\% & -1.03\% & 100.7\% & 100.8\% \\
    \bottomrule
    \end{tabular}%
  }
  \label{table:1}
\end{table*}%

% Table generated by Excel2LaTeX from sheet 'Sheet2'
\begin{table*}[!t]
  \centering
  \caption{Experimental Results of ECM-12.0, Anchor: ECM-12.0 Without CCSAO}
  \resizebox{0.9\linewidth}{!} {
    \begin{tabular}{cccccc|ccccc}
    \toprule
    \multirow{2}[4]{*}{Class} & \multicolumn{5}{c|}{AI}               & \multicolumn{5}{c}{RA} \\
\cmidrule{2-11}          & Y     & Cb     & Cr     & EncT  & DecT  & Y     & Cb     & Cr     & EncT  & DecT \\
    \midrule
    A1    & -0.28\% & -0.83\% & -1.36\% & 104.4\% & 105.0\% &       &       &       &       &  \\
    A2    & 0.01\% & -0.99\% & -1.15\% & 102.2\% & 103.2\% &       &       &       &       &  \\
    B     & 0.08\% & -1.94\% & -1.63\% & 99.0\% & 100.2\% & -0.16\% & -3.76\% & -4.07\% & 102.8\% & 101.5\% \\
    C     & 0.11\% & -0.83\% & -0.41\% & 99.3\% & 100.2\% & 0.00\% & -2.10\% & -1.20\% & 100.4\% & 100.0\% \\
    E     & 0.02\% & -1.55\% & -0.68\% & 100.3\% & 101.7\% &       &       &       &       &  \\
    \midrule
    Average & 0.01\% & -1.28\% & -1.08\% & 100.7\% & 101.7\% &       &       &       &       &  \\
    \midrule
    D     & 0.03\% & -0.02\% & -0.31\% & 98.1\% & 95.9\% & 0.10\% & -1.56\% & -1.05\% & 105.5\% & 104.0\% \\
    F     & -0.23\% & -1.99\% & -1.74\% & 99.4\% & 100.3\% & -0.15\% & -2.99\% & -1.54\% & 105.0\% & 99.9\% \\
    TGM   & -0.73\% & -1.64\% & -1.81\% & 103.8\% & 101.2\% & -1.01\% & -2.72\% & -3.38\% & 107.2\% & 95.3\% \\
    \bottomrule
    \end{tabular}%
    }
  \label{table:2}
\end{table*}%

\if false
\begin{table}[H] %h表示三线表在当前位置插入  
\caption{ECM-12.0 without CCSAO vs. ECM-12.0}
\label{table:2}
    \centering
    \vspace{5pt}
    \centering
    \begin{tabular}{ccccc}
    
    \toprule  %添加表格头部粗线
    Configuration & Class & Y & Cb & Cr\\
    \midrule  %添加表格中横线
    \textbf{AI} & A1 & -0.28\% & -0.83\% & -1.36\%\\
    \textbf{} & A2 & 0.01\% & -0.99\% & -1.15\%\\
    \textbf{} & B & 0.08\% & -1.94\% & -1.63\%\\
    \textbf{} & C & 0.11\% & -0.83\% & -0.41\%\\
    \textbf{} & E & 0.02\% & -1.55\% & -0.68\%\\
    \midrule  %添加表格中横线
    \textbf{} & \pmb{Overall} & \pmb{0.01\%} & \pmb{-1.28\%} & \pmb{-1.08\%}\\
    \midrule  %添加表格中横线
    \textbf{} & D & 0.03\% & -0.02\% & -0.31\%\\
    \textbf{} & F & -0.23\% & -1.99\% & -1.74\%\\
    \textbf{} & TGM & -0.73\% & -1.64\% & -1.81\%\\
    \midrule  %添加表格中横线
    \textbf{RA} & B & -0.16\% & -3.76\% & -4.07\%\\
    \textbf{} & C & 0.00\% & -2.10\% & -1.20\%\\
      \midrule  %添加表格中横线
    \textbf{} & \pmb{Overall} & \pmb{-0.09\%} & \pmb{-3.02\%} & \pmb{-2.79\%}\\
    \midrule  %添加表格中横线
    \textbf{} & D & 0.10\% & -1.56\% & -1.05\%\\
    \textbf{} & F & -0.15\% & -2.99\% & -1.54\%\\
    \textbf{} & TGM & -1.01\% & -2.72\% & -3.38\%\\
    \bottomrule %添加表格底部粗线
    \end{tabular}
\end{table}

\fi
  
\section{Conclusion}

    Cross-component filters play a crucial role in the future of video coding standards.
    By leveraging the correlation between luma and chroma components, cross-component filters can achieve substantial coding performance improvement, leading to the adoption of various video coding standards such as VVC and AVS3.
    Compression distortion can be effectively mitigated, thereby improving the accuracy of the reconstructed pixel.
    Nevertheless, the philosophy of current cross-component filters primarily emphasizes utilizing luma information to refine chroma pixels, which neglects the potential impact of chroma information on luma pixels and the correlation between two chroma components.
    In some scenarios, the chroma texture information and edge details can also contribute to correcting luma inaccuracies. 
    %\jiaqi{Moreover, xxxx. }
    Therefore, cross-component filters still have the potential to achieve substantial performance improvement by delving into the filtering manner and relationship between different channels.

    %Tool is a type of tool that enhances video encoding efficiency by leveraging strong correlations between different components in current video sequences. 
    
    %Typically, it is built upon traditional single-component in-loop filter tools and calculates parameters required for cross-component filtering of the chroma channel based on feature information in the luma channel. 
    
    %This process aims to reduce sample distortion in the chroma component and enhance chroma fidelity. 
    %Experiment results demonstrate the excellent performance of cross-component filtering, leading to its adoption in various video coding standards such as VVC and AVS3.

    %However, current cross-component techniques primarily focus on using information from the luma component to refine chroma reconstruction samples. 
    %In some scenarios, the chroma component may contain necessary feature information for the luma component, such as edges, texture information, etc. 
    %It suggests that there may still be untapped potential to further enhance the correlation between different components in video sequences.

%\begin{refcontext}[sorting = none]
%\bibliographystyle{ieee_fullname}
\bibliographystyle{unsrt}
\bibliography{ref}
%\end{refcontext}
}

\end{multicols}

\end{document}